\documentclass[letterpaper, 10 pt, conference]{ieeeconf}
\IEEEoverridecommandlockouts
\usepackage{epsfig}
\usepackage{graphicx}
\usepackage{subcaption}
\usepackage{amsmath}
\usepackage{amssymb}
\usepackage[utf8]{inputenc}
\usepackage{algorithm}
\usepackage[noend]{algpseudocode}
\usepackage{makecell}
\usepackage[pagebackref=true,breaklinks=true,letterpaper=true,colorlinks,bookmarks=false]{hyperref}

\newcommand{\sfunction}[1]{\textsf{\textsc{#1}}}
\algrenewcommand\algorithmicforall{\textbf{foreach}}
\algrenewcommand\algorithmicindent{.8em}
\floatname{algorithm}{Procedure}

\title{\LARGE \bf CoBe - Coded Beacons for Localization, Object Tracking,\\ and SLAM Augmentation}

\author{Roman Rabinovich$^{1*}$ and Ibrahim Jubran$^{2*}$ and Aaron Wetzler$^1$ and Ron Kimmel$^1$
\thanks{*Equal contribution by authors.}
\thanks{$^1$ Geometric Image Processing (GIP) Lab, Computer Science Department, Technion – Israel Institute of Technology, Israel.}
\thanks{$^2$ Robotics \& Big Data (RBD) Lab, Computer Science Department, University of Haifa, Israel.}
}

\begin{document}
\maketitle
\thispagestyle{empty}
\pagestyle{empty}

\begin{abstract}
This paper presents a novel beacon light coding protocol, which enables fast and accurate identification of the beacons in an image.
The protocol is provably robust to a predefined set of detection and decoding errors, and does not require any synchronization between the beacons themselves and the optical sensor.
A detailed guide is then given for developing an optical tracking and localization system, which is based on the suggested protocol and readily available hardware.
Such a system operates either as a standalone system for recovering the six degrees of freedom of fast moving objects, or integrated with existing SLAM pipelines providing them with error-free and easily identifiable landmarks.
Based on this guide, we implemented a low-cost positional tracking system which can run in real-time on an IoT board.

We evaluate our system's accuracy and compare it to other popular methods which utilize the same optical hardware, in experiments where the ground truth is known.
A companion video containing multiple real-world experiments demonstrates the accuracy, speed, and applicability of the proposed system in a wide range of environments and real-world tasks.
Open source code is provided to encourage further development of low-cost localization systems integrating the suggested technology at its navigation core.

%In this paper we describe a unique beacon light coding method for optical tracking and localization. Our method is robust to bit errors and do not require synchronization. The described method is designed to be used in industrial or consumer environments for full standalone 6dof tracking or as known error free landmarks in a SLAM pipeline. We then present our low cost implementation of such tracking system along with experimental results and use case demonstration.
%We publish the source code an full implementation details for the benefit of the autonomous system community and researchers in other fields

\end{abstract}

\section{Introduction}\label{sec:Intro}

Numerous situations exist in the modern world where accurate knowledge of the position and orientation of an object or person are required in real time ($>20$ frames per second) with $cm$ to sub-$mm$ level accuracy.
This problem is known as positional or spatial localization (includes orientation) and is necessary for many applications including tracking in Virtual Reality, autonomous vehicle and robot navigation, warehouse automation, to name just a few.

Methods for positional localization are generally organized into several categories: inertial, magnetic, wireless, acoustic, external and internal optical tracking.
The inertial approach is cheap and easily scalable, but has large drifts over time~\cite{diaz2019review}. Magnetic tracking is very accurate but is not scalable and is very expensive~\cite{mazuryk1996virtual}. Wireless tracking are either very inaccurate below $100$mm or require additional expensive sensors~\cite{liu2007survey}. Acoustic tracking has low update rates, latency problems, and is sensitive to environment changes such as temperature and humidity~\cite{mazuryk1996virtual}.

%Inertial tracking is an inexpensive method that has high update rate with low latency and is not limited to a specific area but suffer from large drifts over time~\cite{diaz2019review}.
%Magnetic tracking is very accurate, has fast update rates and zero drift but is not scalable to large areas, works poorly near certain objects such as metal and is very expensive~\cite{mazuryk1996virtual}.
%Wireless tracking cannot provide an accuracy below $100$ mm without using sensor fusion or expensive specialized hardware that requires special licensing~\cite{liu2007survey}.
%Acoustic tracking has low update rates and latency problems.
%In addition, they suffer from sensitivity to environment changes such as temperature, atmospheric pressure, and humidity~\cite{mazuryk1996virtual}.

Optical tracking offers a variety of solutions for the localization challenge.
Each solution can overcome different limitations depending on the use case.
These optical tracking systems are generally divided into two approaches, internal and external.
In the internal approach the optical sensors is mounted on the tracked object along with a processing unit.
The sensor perceives the environment and the processor tries to calculate its pose.
This approach allows the tracked object to be independent from external equipment and free from spatial limitations.
This approach might not be suited when there is a limitation on the object's size and weight which cannot allow for the sensor and processor to be mounted.
On the other end, the external approach relies on an optical sensor which is usually in a stationery location along with its processing unit.
The tracked object (or objects) need to be in the sensor's field of view.
This approach allows the tracked objects to be light weight and require minimal (or none) of added equipment on them.
The limitation of this approach is that the tracked objects must be in the view of the sensor, which restricts their operating range, and that in some cases a wireless connection might be required between the external monitoring device and the tracked object to send location-based control commands.

The main challenge in both internal and external approaches is the extraction of features or landmarks that can (i): be easily identified and tracked (in the internal approach, those features are landmarks in the observed scene, and in the external approach its the object itself), and (ii): provide information about the spatial position and orientation of the observer.
Extensive research has been done to address this challenge.
The solutions can be roughly divided into two categories: marker-based and marker-less solutions.
Marker-based methods deploy known objects in the observed scene (external approach) or mount them on the tracked object (internal approach).
Marker-less solutions try to extract (previously unknown) features from its unknown environment.
We will expand on both types of solutions in Section~\ref{sec:Related work}.

In this paper, we propose a novel vision marker-based positional tracking system; see Section~\ref{sec:TrackingSystem}.
The system is based on a novel coded-light method which will be described in Section~\ref{sec:coded-light}.

%________________________________________________
\section{Related work}\label{sec:Related work}
In this section we describe the main related optical-based methods that have been used for positional tracking in recent years.

In fields such as mobile robotics, it is common to use internal tracking solutions (ego-motion); see Section~\ref{sec:Intro}.
For this goal, many marker-based systems were suggested which utilize known markers for localization, e.g., a chessboard, ARTag~\cite{fiala2005artag}, arUco~\cite{garrido2014automatic}, and~\cite{bruckstein2000new}, which are also known as fiducial markers.
Generally, such techniques rely on the known geometry of these patterns, which are distributed around the working environment in advance;
These techniques are susceptible to illumination variations and partial occlusions, they do not scale up to larger environments due to the need for close visual views of the patterns (see Section~\ref{sec:Experiments}), and require the ability to distinguish between the different detected patterns.

To this end, an alternative marker-less technique, called simultaneous localization and mapping (SLAM), is widely used~\cite{taketomi2017visual,mur2015orb}.
%This technique uses a digital sensor coupled with a processing unit which detects, identifies, and tracks 2D image points in order both to estimate the sensor motion and to reconstruct, in real-time, the structure of the unknown environment; see e.g.,~\cite{mur2015orb}.
SLAM systems do not usually require prior knowledge and make use of readily available camera technologies.
However, SLAM systems suffer from issues of drift, relocalization (see Section~\ref{sec:Experiments}), as well as sensitivity to environment variation due to its dependency on natural features.
In addition, those methods usually require a strong processor since they apply computationally heavy heuristics, such as bundle adjustment~\cite{triggs1999bundle}, to compensate for their inaccuracies and uncertainties.
To overcome the drift and relocalization problems, hybrid systems were developed which also utilize predefined markers in this SLAM pipeline, e.g.~\cite{munoz2020ucoslam} which use arUco markers to improve robustness.
However, these hybrid systems still suffer from the same problems as the marker based systems described above.
Another internal localization system, which is based on active markers, is the HTC VIVE lighthouse~\cite{HTCVIVE}, mainly used for virtual environment systems and indoor localization; see~\cite{niehorster2017accuracy}.
%The lighthouse is an active marker that includes a rotary motor that works in high speeds and IR lighting.
%One or more of these devices are spread in the working environment, depending on the desired area size.
%A mobile tracker, which consists of several IR receivers deployed in a specific formation, receives the IR light that is transmitted from the lighthouse and is able to calculate its own pose.
This system can provide a precise pose information at high speeds. However, it is limited to a small room scale due to the need to synchronize the lighthouse(s) with the IR receivers deployed on the object. It also includes sophisticated and costly moving mechanical parts which are sensitive to environment noise and are more prone for failure.

The external optical tracking approached is used when there are limitations on the payload that the object can carry, or when multiple objects need to be tracked in the same reference frame.
OptiTrack~\cite{OptiTrack} and the Vicon~\cite{Vicon} systems are arguably amongst the most popular commercial marker-based systems which apply the external approach.
These systems can produce high precision positional data at high frame rates but are expensive and require a long calibration procedure and high computational power. Some alternative low-cost systems, which use different variations of known markers, were suggested in~\cite{faigl2013low,nasser2015low}.
These are usually non-robust due to their dependency on the 2D view of the predefined geometric pattern placed on the tracked object.

%Here, we propose and implement a novel positional tracking pipeline which is simultaneously robust to illumination changes and challenging environments, easy to calibrate and deploy, and operates in real-time using cheap off-the-shelf hardware.

%________________________________________________
\section{Our contribution and novelty} \label{sec:contrib}
Our paper introduces the following contributions:
\begin{enumerate}
\renewcommand{\labelenumi}{(\roman{enumi})}
\item We provide a code-book containing code-words which are provably error resistant and have an efficient decoding process.

\item A novel coded beacons (markers) which are easy to extract and identify, based on our code-book.
These markers, which we call CoBe (coded beacons), flash a unique cyclic binary code at high speeds, and do not require a shared clock signal for synchronization; see Section~\ref{sec:coded-light} and patent~\cite{wetzler2018tracking}.

\item Based on our CoBe, we propose a general pipeline for a positional tracking system which can function in both internal and external approaches; see Section~\ref{sec:sysOverview}.
We then implement our own low-cost off-the-shelf infra-red system based on this pipeline; see Section~\ref{subsec:implementation}.

\item Extensive real-world tests demonstrate our system's practical applications, real-time performance, and accuracy; see Section~\ref{sec:Experiments}.

\item A full open source code for our system along with a simple guide for its reproduction~\cite{opencode}. We hope such an open source code and guide will lead to the development of more homemade tracking systems based on the proposed technology.
\end{enumerate}

%________________________________________________
\section{Coded light codes} \label{sec:coded-light}
In this section we describe our key contribution: a coded light sequences scheme that can be used as robust location identifiers.
We start by briefly describing the physical transmitter properties that we leverage to generate the coded sequences.
We then describe a simple coded light protocol to demonstrate the main concept, followed by the more involved one we used.
The protocol was first patented in~\cite{wetzler2018tracking} and  is now provided as an open source code~\cite{opencode}.

\subsection{Transmitter properties} \label{sec:transProp}
We use optical beacons that emit a cyclic binary sequence of light.
This can be done either with a low-high brightness sequence using near-visible (infra-red) illumination or a red-blue visible light sequence, depending on the specific use and environment.
The flashing light sequence is interpreted as a binary code by extracting the bit value at each time interval that corresponds to single bit in the sequence.
The bit value extraction method we used is detailed in Section~\ref{subsec:implementation}.
%More extraction options are given in the supplementary material; see Section~\ref{}.

\subsection{Robust cyclic binary codes for unique identity detection}
The use of coded light pulses for identifying light beacons has a long history in the field of navigation, manifesting itself in lighthouses and way-point buoys among others; see for example \cite{lightlist}.
We wish to continuously transmit a binary light code that can be used to uniquely identify a beacon.

In order to make the coded-light scheme robust and useful for an automated detection and decoding pipeline, we wish to design a code-book containing code-words that satisfy the following three properties:
%Because navigational codes are designed to be observed and decoded by human operators they span timescales of seconds or minutes. This enables every code bit to be continuously observed and timed by a person or alternatively sampled numerous times by electronic sensors. In direct contrast to this scenario we describe a coding system for which it is assumed that only a single sample of every bit will occur and the sampling clock of the optical sensor is not synchronized with the beacon emission clock.
%Our contribution here is therefore to specify code-words that are designed to have three properties which make them robust, and useful for an automated detection and decoding pipeline:
\begin{itemize}
  \item fast lock-on and decoding time that is close to a single full code cycle,
  \item robustness to single bit-shift and bit-flip errors,
  \item code-book size at least linear in code-word length.
\end{itemize}
It is important to note that realizing only one or two of these properties without the other is considerably simpler than satisfying all three together.

%___________________
\subsubsection{\bf Initial approach - no error robustness}
{\label{subsec:InitialApproach}}
To design a code-book with code-words that can be detected within one full code cycle without the need to use a synchronization comma we initially choose the set $C_n$ of complete cyclic equivalence classes \cite{pearson2003comma} using n-bit binary codes.
Every element $c \in C_n$  is defined as the set of the $n$ cyclic shifts of the $n$-bit sequence $(b_1,b_2,...,b_n)$.
In other words $c$ has the property that if $(b_1,b_2,...,b_n) \in c$ then $(b_i,b_{i+1},...,b_n,b_1,...b_{i-1}) \in c$ for every $i \in \{1,\cdots,n\}$.
We define $c_0$ to be the number in $c$ whose decimal value is the smallest.
We denote by $c_i$ the code-word in $c$ with a cyclic shift left of $i$ bits relative to $c_0$.
%Since $c_0,\cdots,c_n \in c$ are all equivalent up to cyclic bit-shifting,
Let $c_0$ be the representative of the set $c$ and consider its decimal value to be the representative identifier $id$.
These codes are particularly useful in our case because the same code-word is  continuously transmitted by a specific beacon.
Assuming there are no errors and that the optical sensor receives at least $n$ consecutive bits from a given beacon, we can unambiguously decode (identify) the code-word irrespective of the observation starting point in the bit stream.
The most trivial cyclic codes one could think of are $(00\ldots 0)$ and $(11 \ldots 1)$, and in our case are not useful, so we ignore them.
An example of a single element in $C_4$ is the set $c = \{(0001),(0010),(0100),(1000)\}$.
The total number of codes available in $C_n$ is exponential in code-word length.
% is given by Integer Sequence A000031~\cite{}.

\paragraph{\bf Creating the code-book}
In order to construct the code-book we need to find all the complete cyclic equivalence classes.
We store a list of the $c_0$ code-words for every such class $c$, for example, via simple exhaustive search.
We also maintain a lookup table that stores every possible $n$-bit sequence $c_0,\cdots,c_n$ together with the corresponding $c_0$ code-word from which it can be generated.
%The algorithm to generate these data structures is presented in Algorithm \ref{alg:codebookinitial} in the appendix. Section~\ref{sec:includeNoise} presents an improved version of that algorithm.
%In practice, we map the trivial code-words with all $1$s or $0$s to $0$ in the lookup table.

\paragraph{\bf Encoding}
We choose an identifier from $0$ to $|C_n|-1$ to represent a beacon.
To encode this identifier we use the identifier as an index into the code-book and take the $c_0$ stored at that location.
%This is shown in Algorithm \ref{alg:encodeinitial}.
The process of transmission is now trivial.
We transfer the code to a beacon and periodically transmit the bits at a known bit rate using either the intensity or hue based methods described in Section~\ref{sec:transProp}.
%
%\begin{algorithm}
%\caption{\label{alg:encodeinitial} Encode ID}
%\begin{algorithmic}[1]
% \Require $\mathit{C_n},id$ \Comment code-book, code number
% \Ensure $c_0$
% \State $c_0 \gets \mathit{C_n}(\mathit{id})$
%\end{algorithmic}
%\end{algorithm}

\paragraph{\bf Decoding}
Once the optical sensor has detected and recorded at least $n$ consecutive bits from a given beacon,  decoding the received code-word to obtain the beacon identity is performed.
Only the last $n$ bits are processed simply by accessing a (pre-computed) look-up table that maps the bit sequence into a representative beacon identity.
% Which is a computationally efficient decoding.

%If for some reason we retrieve a $0$ as the identifier then it means the detected bit sequence is unknown.
%It is instructive to notice that only $n=4$ bits were required until the decoder could lock-on and determine the identifier.
%Furthermore, the decoding process is just a table look-up which implies fast decoding provided that the processing unit's memory is quickly accessible or the table is implemented in hardware.

\begin{table}[htbp]
\centering
\begin{tabular}{c c}
{%
\begin{tabular}{@{}l|llll@{}}

$c_0$ & 0001 & 0011 & 0101 & 0111 \\
$id$ & 1 & 3 & 5 & 7 \\
\end{tabular}
}
\\
{%
\begin{tabular}{@{}ll | ll | ll | ll@{}}
\hline
code & $id$ & code & $id$ & code & $id$ & code & $id$ \\
\hline
0000 & 0 & 0100 & 1 & 1000 & 1 & 1100 & 3 \\
0001 & 1 & 0101 & 5 & 1001 & 3 & 1101 & 7 \\
0010 & 1 & 0110 & 3 & 1010 & 5 & 1110 & 7 \\
0011 & 3 & 0111 & 7 & 1011 & 7 & 1111 & 0 \\
\hline
\end{tabular}
}
\end{tabular}
\caption{\label{tab:C_4} $C_4$ code-book and $D_4$ look-up table mapping every code-word to its decimal value identifier in the non-robust manner.}
\end{table}

As an example, a look-up table for $C_4$ can be seen in Table~\ref{tab:C_4}.
If the following sequence was received (left to right) $1101110111$, it would have been decoded in consecutive subsets of $n=4$ bits as $(1101) \to 7$, $(1011) \to 7$, $(0111) \to 7$ and so on.
The resulting decoding stream would be $7777777$ as expected.

%\begin{algorithm}
%\caption{\label{alg:decodeinitial} Decode n-bit code}
%\begin{algorithmic}[1]
% \Require $\mathit{D_n},c_i$ \Comment look-up table, binary sequence % \Ensure $c_0$
% \State $ \mathit{id} \gets \sfunction{Bin2Dec}(c_i)$
% \State $c_0 \gets \mathit{D_n}(\mathit{id}) - 1$
%\end{algorithmic}
%\end{algorithm}

%_________________________________________
\subsubsection{\bf Robustness to noise} \label{sec:includeNoise}
The approach to generate and decode binary codes described above is straightforward and over-simplistic.
However, it lacks any ability to detect and correct errors.
Typical error correcting approaches encode bit sequences with parity bits, Hamming codes, CRC codes and others.
These methods measure the distance between two code words using the Hamming metric, and ensure a minimal Hamming distance between all code-words.
Our experiments demonstrated that we should consider the three following types of bit errors that can occur with the same probability,
(i) bit-flip, where a bit is interpreted incorrectly ($1$ as $0$ or vice versa),
(ii) bit-miss, where a bit was not received, and
(iii) bit-insertion, where an additional bit was falsely added to the sequence.

%The Hamming metric simply counts the number of bits which are different between two code-words, i.e., bit-flips.
%Unfortunately, when also bit-misses and bit-insertions occur, the Hamming distance is insufficient. A considerably more appropriate metric is the Levenshtein distance \cite{Levenshtein}. This measures how many changes addition/deletion/flip operations on one of the sequences. A bit-flip error would result in a Levenshtein distance of 1. A bit-miss and bit-insertion errors are considered as a chained insertion and deletion operations (or vice-versa) on the received sequence and has a Levenshtein distance of 2, since we always compare the last $n$ seen bits to some code of length also $n$. However, two strings from the same cyclic equivalence class (the same string up to some cyclic shift), which basically represent the same string, may have a very large Levenshtein distance. So even though the Levenshtein distance is more appropriate for error detection we will not use it directly.

It is common to consider the Hamming or the Levenshtein distance~\cite{Levenshtein} as the error metric between two bit sequences.
However, a closer look reveals that these metrics may yield a large distance for two strings that belong to the same cyclic equivalence class (the same string up to some cyclic shift).

\begin{figure}[htbp]
    \centering
    \begin{subfigure}[t]{.49\columnwidth}
        \includegraphics[width=\linewidth]{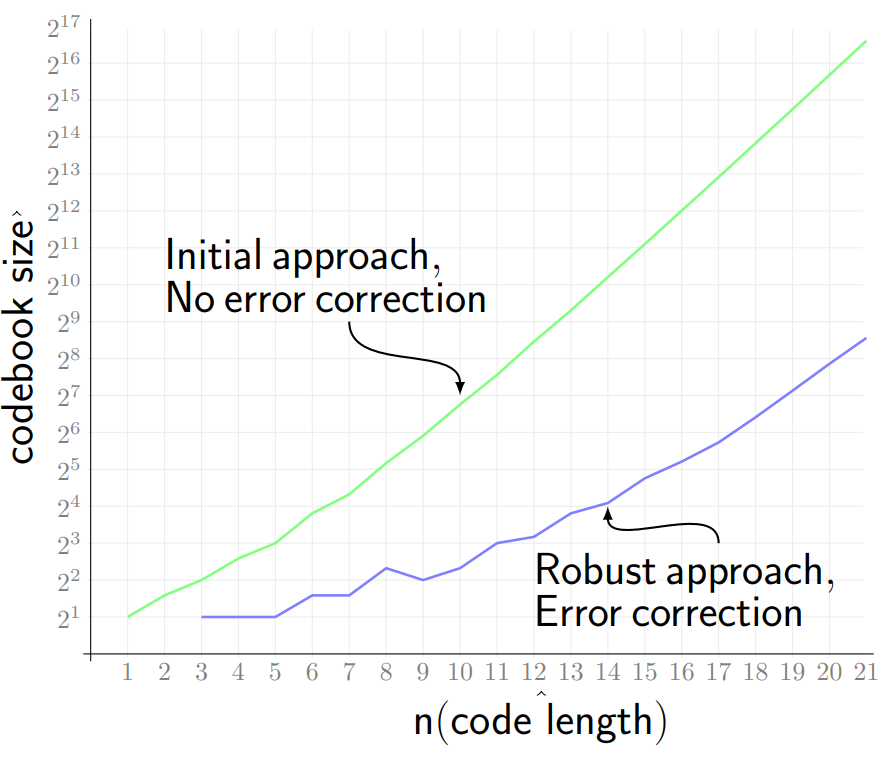}
        \caption{Comparison of code-book sizes for different code-word lengths for the initial and robust code-book generation methods.}
        \label{fig:numcodes}
    \end{subfigure}
    \begin{subfigure}[t]{.49\columnwidth}
        \includegraphics[width=\linewidth]{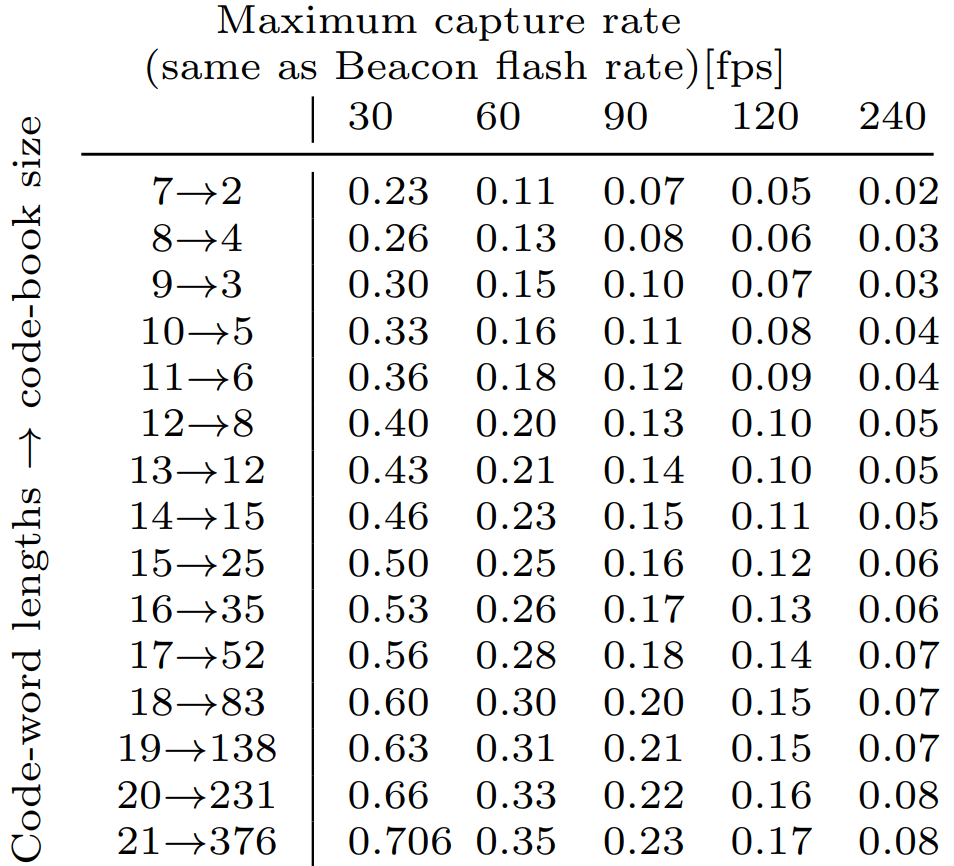}
        \caption{Table of lock-on times in seconds for code-book size versus sample frame-rate.}
        \label{tab:tradeoff}
    \end{subfigure}
\end{figure}

Instead, in order to construct a robust code-book for every cyclic equivalence class $c$, we would like to find all code-words that can be obtained from some $c_i \in c$ by exactly one error from the above permissible errors, and map all these code-words via our look-up table to the same identifier (representative) $c_0$ of $c$.
%To find the maximal set of non-overlapping classes as explained above, we initially used the Bron-Kerbosch~\cite{bron1973algorithm} algorithm. However, solving the maximal independent set problem is NP-complete and finding a solution became unfeasible for $n$ larger than $14$. Instead,
We found that a simple greedy algorithm was sufficient to compute the maximal set of non-overlapping classes as explained above and produce a code-book with size exponential in the number of bits; see Fig.~\ref{fig:numcodes}.
Besides the cyclic variations (equivalences) of $c_0$, the suggested greedy algorithm also enumerates and stores all error variants of these cyclic variations in a look-up table; see Algorithm~\ref{alg:codebookrobust}.

% This greedy algorithm is almost exactly the same the error-less Algorithm described above. However, instead of only storing the cyclic variations (equivalences) of $c_0$ in the look-up table, we also enumerate and store all the error variants of all these cyclic variations in the look-up table; see Algorithm~\ref{alg:codebookrobust}.
% The code-books generated by this procedure can be seen in the supplementary material?? and
The number of codes produced for both this method and the non-robust initial method can be seen in Fig.~\ref{fig:numcodes}.
{\bf Encoding} and {\bf Decoding} are exactly as explained in the non-robust approach suggested above.

% \figtex{graphs/numcode.tex}{Comparison of code-book sizes for different code-word lengths for the initial and robust code-book generation methods.}{fig:numcodes}{}

\begin{algorithm}
\caption{\label{alg:codebookrobust} Generating robust $n$-bit cyclic equivalence classes}
\begin{algorithmic}[1]
 \Require $n$ \Comment $n$ code bits
 \Ensure $\mathit{D_n},\mathit{C_n}$ \Comment look-up table, code-book
 \State $\mathit{D_n} \gets \sfunction{zeros}(1,2^n)$ \Comment look-up table
 \State $\mathit{C_n} \gets \emptyset$ \Comment code-book
 \State $\mathit{k} \gets 0$ \Comment code counter(identifier)

\ForAll{$c_0\in\mathit{(0,..,2^n-1) }$} \Comment $c_0$ is decimal
    \State $\mathit{inC} \gets \mathit{false}$
    \State $\hat {c_0} \gets \sfunction{Dec2Bin}(c_0, n)$ \Comment $\hat {c_0}$ is n-bit binary
    \State $\mathit{Id} \gets \emptyset$ \Comment look-up table set of indices

    \ForAll{$i\in\mathit{(0,..,n-1) }$}
        \State $\hat {c_i} \gets \sfunction{CyclicShift}(\hat {c_0},i)$
        \State $\mathit{Idx} \gets \{\mathit{Idx},\sfunction{Bin2Dec}(\hat {c_i})\}$
        \State $\mathit{V} \gets \sfunction{Noisify}(\hat {c_0})$ \Comment insertions, deletion, flips

        \ForAll{${\hat v}\in\mathit{V}$}
            \State $\mathit{id} \gets \sfunction{Bin2Dec}(\hat v)$
            \State $\mathit{Id} \gets \{\mathit{Id,id}\}$
            \State $\mathit{inC} \gets (\mathit{inC}) \vee (\mathit{D_n}(\mathit{idx}+1)>0)$
        \EndFor
     \EndFor
     \If{$\neg \mathit{inC} $} \Comment if $c_0$ not overlaps with $\mathit{C_n}$
        \State $\mathit{k} \gets \mathit{k} + 1$ \Comment increase code counter
        \State $\mathit{C_n} \gets \{\mathit{C_n} , c_0\}$ \Comment add $c_0$ to code-book
        \ForAll{$\mathit{id} \in \mathit{Id}$} \Comment for all noised $c_i$ ...
            \State $\mathit{D_n}(\mathit{id}) \gets k$ \Comment ... stores $\mathit{k}$ in $\mathit{D_n}$
        \EndFor
     \EndIf
 \EndFor
\end{algorithmic}
\end{algorithm}

\begin{table}[htbp]
\centering

\begin{tabular}{c c}
{%
\begin{tabular}{ c | c | c | c }
\hline
$c_0$ & 00000111 & 00111111 & 01010101 \\
$id$ & 7 & 63 & 85\\
\end{tabular}
\vspace{1pt}
}
\\
{%
\begin{tabular}{@{}c | c | l@{}}
\hline
code & mapped $id$ & \quad\quad\quad\quad\quad error \\
\hline
00010100 & 7 & 2 bit cyclic shift + bit-flip\\
00001101 & 7 & bit-insertion\\
00101111 & 63 & bit-flip\\
00110111 & 63 & bit-insertion\\
00101010 & 85 & bit-insertion\\
10100101 & 85 & cyclic shift + bit-miss\\
01100010 & unmapped & \makecell{one error away from a variant of id:7}
\end{tabular}
\vspace{5pt}
}
\end{tabular}
\caption{ \label{tab:C_5_robust}
$C_5$ code-book and $D_5$ lookup table, built using the robust method.
The final table is a union over the error variants.
For example $11110$ is a bit-flip error of $11100$.
 Similarly $01101$  bit-shift error.}
\end{table}

%_______________________________________________________________
\section{Positional Tracking System} \label{sec:TrackingSystem}
In this section, we specify both the hardware and the software procedures required to construct and operate a full six degrees of freedom (6DoF) positional tracking system which is based on the coded-light scheme presented in Section~\ref{sec:coded-light}.
We then demonstrate in detail our low-cost implementation for such a system in Section~\ref{subsec:implementation}.
The system developed in this paper is used in both the internal and external vision-based approaches described in Section~\ref{sec:Intro}.
We demonstrate in Section~\ref{sec:Experiments} our system's applicability, performance, and accuracy in real-world tasks.

%_______________________________________________________________
\subsection{System overview}
\label{sec:sysOverview}

\paragraph{\bf Hardware}
The system consists of two main components.
The first component, called the \emph{perception unit}, requires an optical sensor (for example a camera) and a processing unit.
The second component, denoted as the \emph{marker unit}, contains a set of \emph{beacons}.
A beacon is a device equipped with a micro-controller connected to some light source (for example a LED).
Each beacon has a unique code based on the coded-light protocol presented in Section~\ref{sec:coded-light}.
The micro-controller transmits this unique binary code by alternating between the two states of the light source.

\paragraph{\bf Calibration and mapping}
The following is a pre-processing step that needs to be carried out only once, before activating the system.
The optical sensor needs to be calibrated in advance to obtain its internal parameters (e.g. the intrinsic parameters in the case of a simple camera). This can be done using one of the many tools available online, e.g., OpenCV~\cite{opencv_library}. For more details on internal optical sensor calibration see~\cite{zhang2000flexible}.

Now, the beacons need to be deployed as follows: in the internal approach, the beacons should be spread around the desired working environment.
In the external approach, the beacons should be placed on the object(s) to be tracked; see Fig.~\ref{Fig:BeaconDeploy}.

\iffalse
\begin{figure}[htbp]
    \centering
    \begin{subfigure}[t]{.45\columnwidth}
        \includegraphics[width=\linewidth]{images/OfficeEnv1.png}
        \caption{}
    \end{subfigure}
    \begin{subfigure}[t]{.45\columnwidth}
        \includegraphics[width=\linewidth]{images/RaceCar.png}
        \label{fig:car}
        \caption{}
    \end{subfigure}
    \caption{\textbf{Beacon deployment.} (a) Internal tracking: beacons are deployed around an office environment. (b) External tracking: beacons are deployed on a toy car.}\label{Fig:BeaconDeploy}
\end{figure}
\fi
\begin{figure}[htbp]
\centering
\includegraphics[width=\columnwidth]{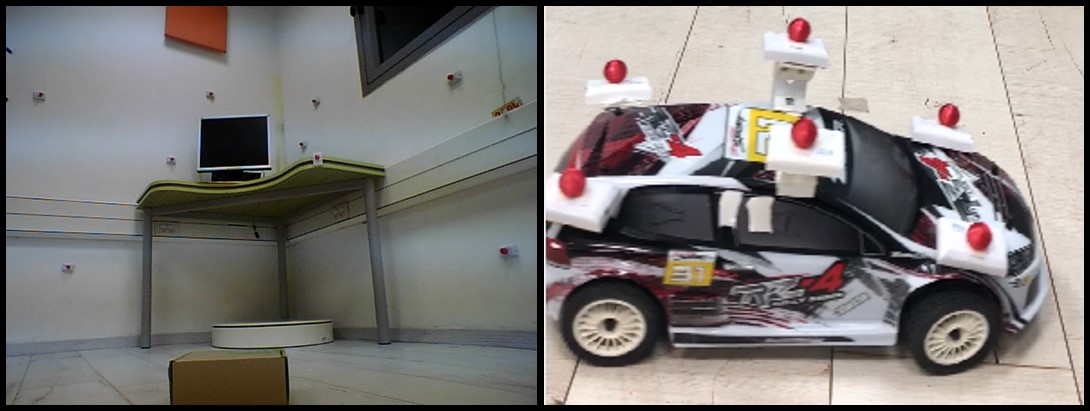}
\caption{\textbf{Beacon deployment.} (Left) Internal tracking: beacons are deployed around an office environment. (Right) External tracking: beacons are deployed on a toy car.}
\label{Fig:BeaconDeploy}
\end{figure}

After the deployment, the relative position of each beacon in 3D space needs to be extracted.
This can be done either by using a structure from motion (SfM) pipeline~\cite{ozyecsil2017survey} (see Section~\ref{subsec:implementation} for a suggested implementation), by using a depth camera, or by manually measuring the pairwise distances between every pair of beacons and then applying the well known multi-dimensional scaling method~\cite{borg2005modern} to embed these beacons in 3D space.
This relative position of the beacons, along with the internal calibration data, are then stored in the processing unit, where all the calculations take part.

\paragraph{\bf Beacon detection and decoding}
Once the prepossessing step is complete, the processing unit can start reading a live stream of frames from the optical sensor, and apply the main procedures: detection and decoding.
In the detection procedure, the goal is to detect the 2D location (region) of each of the visible beacons in the current received frame.
Since a light source is to be detected, the detection can be done by searching for the brightest regions in the frame; see suggested implementation in Section~\ref{subsec:implementation}.

After detecting a set of regions, we associate each of them with their corresponding regions from the previous frame.
Once a beacon was detected for a sufficient number of consecutive frames, the perception unit can identify the unique identity of this beacon by decoding its sequence of states in those frames, and retrieve its 3D location from the previously computed list of beacon 3D locations.

\paragraph{\bf Positional tracking}
If more than $4$ non-planar beacons are successfully identified (decoded) in the same frame, a pose estimation procedure can be invoked to compute the desired 6DoF.
If less than four beacons are detected or those detected are co-planar, then the perception unit may still be able to determine partial degrees of freedom.
Alternatively, the 3D location of the beacons can be utilized to refine and enhance different localization systems running in parallel, such as a SLAM system.

%________________________________________
\subsection{Design considerations}
\paragraph{\bf Beacons distribution}
%The specific distribution of beacons in an environment is done so as to effectively utilize the number of available beacon IDs with code-words of length $n$ that can be generated.
To effectively utilize the number of available beacon IDs, some IDs can be reused in different rooms which are visually separated from each other. For simplicity, we filter out cases where two beacons with the same ID are identified in the same frame.

\paragraph{\bf Frame rates}
The time to first decoding of the beacon ID is called the ``lock-on time''
After the lock-on time, we can identify the beacon after every new bit with no additional time penalty.
The rate at which the beacon emits its code bits is designed to match the rate at which the optical sensor samples the scene.
So, the trade-off is clear: we need high frame-rates for both fast lock-on time, as well as longer code-words while we prefer lower frame-rates to limit the processing requirements of the tracking unit.
This trade-off can be seen in Table \ref{tab:tradeoff}.

\subsection{System implementation} \label{subsec:implementation}
In this section, we discuss in detail our implementation of a positional tracking system based on the overview in Section~\ref{sec:sysOverview}.
Our system was built out of low-cost off-the-shelf components with easy reproduction and accessibility in mind; see illustration in our video~\cite{vid}.
All the code was written in Python using OpenCV~\cite{opencv_library}.
The code is available at~\cite{opencode}.

\paragraph{\bf Coded-light protocol} We used the intensity based method with infra-red (IR) illumination.
The codebook for the coded-light protocol we used was generated using Procedure~\ref{alg:codebookrobust} with $n=15$.

\begin{figure}[htbp]
\centering
\includegraphics[width=\columnwidth]{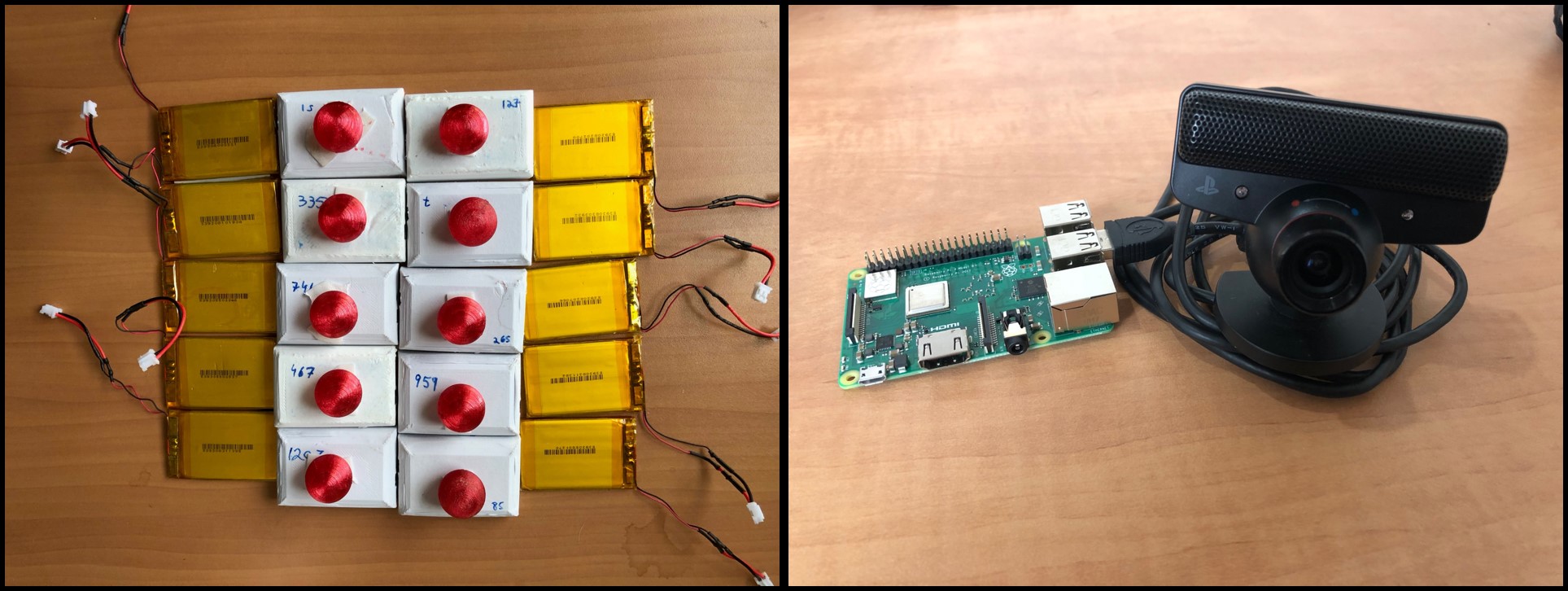}
\caption{\textbf{System components.} (Left) our homemade CoBe (coded beacons) consisting of a microcontroller, a LiPo battery, and an IR LED.  (Right) An IoT board and a cheap webcam; see Section~\ref{subsec:implementation}.}
\label{Fig:systemComps}
\end{figure}

\paragraph{\bf Hardware}
The perception unit consists of a PSEye3 cheap ($<10\$$) webcam, with an IR pass filter attached to the lens.
The camera is connected to a Raspberry Pi~\cite{raspberry}.
The microcomputer can be replaced by a standard laptop.
Each beacon in the marker unit was built using a standard IR LED ($950$nm wavelength) connected to a tiny microcontroller called ``A-star $328PB$ Micro'' with a small LiPo battery, and was assigned a unique code from our chosen codebook.
The microcontroller was programmed to cyclically flash the LED according to this code, where ``1'' and ``0'' in the code correspond to high and low intensity values of the LED, respectively.
To distribute the light evenly in all directions a diffuser cover is placed on each LED; see Fig.~\ref{Fig:systemComps}.
The beacons do not have a shared clock signal.

\paragraph{\bf Calibration and mapping.}
To retrieve the camera's intrinsic parameters, we applied the standard camera calibration pipeline~\cite{zhang2000flexible} using functions from the OpenCV library~\cite{opencv_library}.

We implemented our own simple SfM pipeline which also uses our detection and decoding algorithm for feature extraction and matching between frames. This pipeline applies the following steps: \textbf{(i) Capturing: }capture $m$ frames $F_1,\cdots,F_m$, such that every beacon appears in at least $2$ different frames, and every pair of consecutive frames share at least $5$ common beacons. Each beacon was identified using our decoding scheme detailed in the next paragraph. \textbf{(ii) Relative frame alignment: }compute the essential matrix $E_i$ between every pair $F_i$ and $F_{i+1}$ of consecutive frames as described in~\cite{nister2004efficient}(which is why $5$ common markers are required between the frames), and decompose it into a rotation matrix $R_i$ and translation vector $t_i$ as in~\cite{tsai1984uniqueness}. \textbf{(iii) Triangulation: }triangulate the common points between $F_i$ and $F_{i+1}$ to obtain a point cloud $P_i$. \textbf{(iv) Point-cloud alignment: }align the $m$ point clouds together by transforming them to a common coordinates system, to obtain a combined point cloud $P$ that contains the 3D locations of all the beacons. The optimal translation between two points clouds is simply the vector connecting their means, and the optimal rotation~\cite{wahba1965least} can be simply computed via SVD~\cite{kabsch1976solution}.
\textbf{(v) Refinement via bundle adjustment: }Apply a non-linear least squares optimization function which takes as an initialization the point cloud $P$ and the rough estimation $R_i,t_i$ of the camera pose in every frame $F_i$, and refines $P$ and the poses as to minimize the reprojection error~\cite{triggs1999bundle,more1978levenberg}.

%\paragraph{Calibration and mapping}
%To retrieve the camera's intrinsic parameters, we applied the standard camera calibration pipeline~\cite{zhang2000flexible} using functions from the OpenCV library~\cite{opencv_library}.
%
%We implemented our own simple SfM pipeline which utilizes our detection and decoding algorithm for feature extraction and matching between frames.
%It applies the following steps:
%\textbf{(i) Capturing: } Capture $m$ frames $F_1,\cdots,F_m$, each sharing at least $5$ common beacons, and every beacon appears in at least $2$ different frames.
%Beacon identification is detailed in the next paragraph.
%\textbf{(ii) Relative frame alignment: } Compute the relative rotation and translation between every pair $F_i$ and $F_{i+1}$ of consecutive frames using the Essential Matrix $E_i$~\cite{nister2004efficient}.
%\textbf{(iii) Triangulation: } Triangulate the common points between $F_i$ and $F_{i+1}$ to obtain a point cloud $P_i$.
%\textbf{(iv) Point-cloud alignment: } Align the $m$ point clouds together to obtain one point cloud $P$.
%The optimal translation between two points clouds is simply the vector connecting their means, and the optimal rotation~\cite{wahba1965least} can be simply computed via SVD~\cite{kabsch1976solution}.
%\textbf{(v) Refinement via bundle adjustment:} Refine the point cloud and poses using a non-linear least squares optimization function~\cite{triggs1999bundle,more1978levenberg}.

\begin{figure}[htbp]
    \centering
    \begin{subfigure}{.32\columnwidth}
        \includegraphics[width=\linewidth]{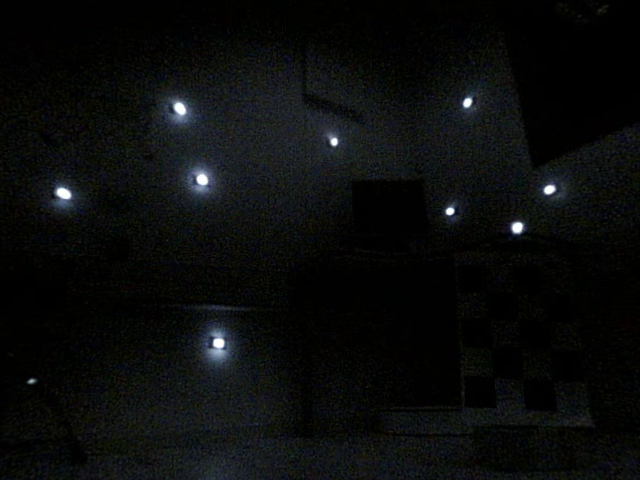}
        \caption{}
    \end{subfigure}
    \begin{subfigure}{.32\columnwidth}
        \includegraphics[width=\linewidth]{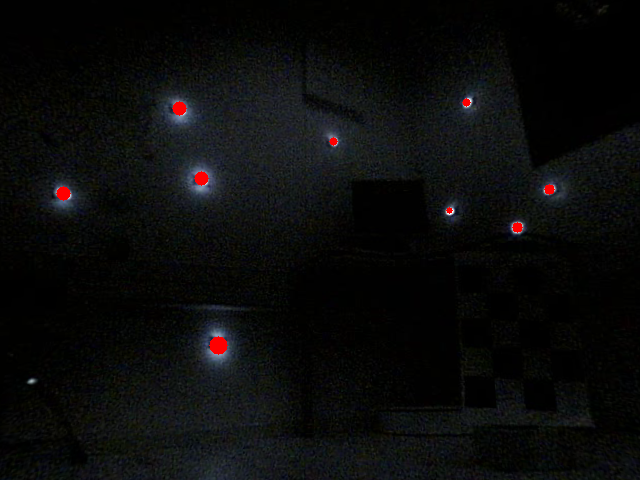}
        \caption{}
    \end{subfigure}
    \begin{subfigure}{.32\columnwidth}
        \includegraphics[width=\linewidth]{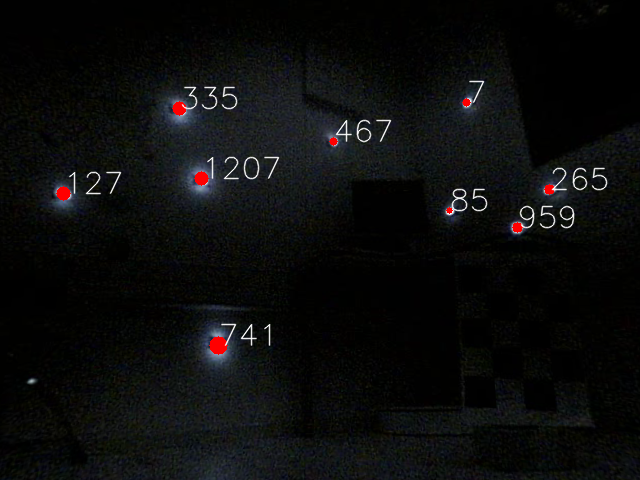}
        \caption{}
    \end{subfigure}
    \caption{beacon detection and decoding: (a)~image received from the sensor, (b)~beacons are detected and tracked (c)~beacons are decoded after n consecutive frames.}\label{Fig:leds}
\end{figure}

\paragraph{\bf Beacon detection and decoding}
Since our camera has an IR pass filter and the beacons use IR LEDS, the beacons appear as bold bright blobs on an (almost) black background, which makes them easy to detect.
Therefore, in each frame $F_i$ we detect the beacons using a simple contour detection function in OpenCV; see Fig. \ref{Fig:leds}.
We also store the area of each contour, which will be used to determine the bit value of the beacon (high or low).
We associate the contours in $F_i$ with a contours from the previous frame $F_{i-1}$ using an optimal matching scheme~\cite{munkres1957algorithms}, which associates the contours so as to minimize their sum of distances in the 2D frames.
We also allow partial matching, that is, associating only a subset of the contours. This is done by increasingly adding ``dummy contours'' and checking the decrease in the final matching cost.

Once every beacon's contour was successfully associated for $n=15$ consecutive frames, we estimate its bit value in each of these frames as follows.
We take the beacon's two contours with the maximal and minimal areas, average the areas, and define the bit value of the beacon as high or low in some frame if its contour area in that frame is larger or smaller than this average, respectively.
To justify this identification, we note that each code has at least one high and one low bit value.
The decoded $15$ bit values correspond to the beacon's desired unique cyclic code.

\paragraph{\bf Positional tracking} If $4$ or more beacon codes have been decoded for some frame $F_i$, we plug their 2D locations (the center of each contour in $F_i$) and their corresponding 3D locations (computed in the pre-processing step) into the function \texttt{SolvePnP} from OpenCV.
We thus obtain the 6DoF of the camera relative to the observed object(s) / environment.

\section{Real-World Experiments} \label{sec:Experiments}
In this section we put the system detailed in Section~\ref{subsec:implementation} to the test.

We conduct different types of experiments which demonstrate the:
(i) Accuracy of our system as compared to other common methods,
(ii) ability of our system to track and localize very fast moving objects and fast lock on and re-localize when occlusions occur,
and (iii) applicability of our system to a wide range of applications and challenging environments, both as a standalone system or as a component fused into existing systems.
The experiments are presented in~\cite{vid}.

\subsection{Accuracy} \label{subsec:Accuracy}
In this experiment, we tested the tracking accuracy of our system when used for the internal tracking approach.
We expect to reach at least the same accuracy as that of the external approach due to the static position of the camera.

\begin{figure}[htbp]
\centering
\includegraphics[width=\columnwidth]{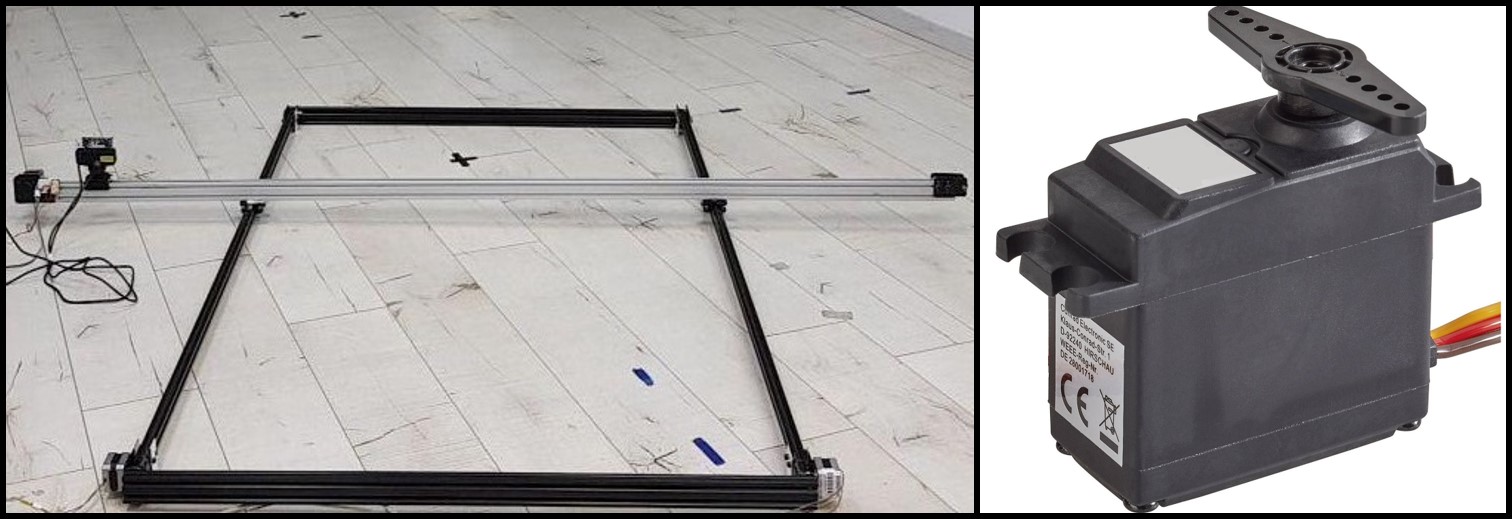}
\caption{\textbf{Systems used as aground truth in our accuracy testing experiments. }(Left) Two axes rail system. (Right) A servo.}
\label{fig:rail}
\end{figure}

\paragraph{\bf Rails test} We mounted our perception unit (camera) on a precise 2-axis moving rail system shown in Fig.~\ref{fig:rail}, which provided the exact ground truth position of the camera, up to the sub-millimeter accuracy. We programmed the two axis rail system to move the camera in a rectangular shape of size $1.2$m$\times 1.2$m, while the exact position of the camera when capturing every frame was known.
The experiment is presented in our video~(at 00:10).

\paragraph{\bf Servo test} To test the accuracy of our system when the camera rotates, we mounted the perception unit on a rotating servo with angular control; see Fig~\ref{fig:rail}.
The servo was programmed to rotate $12$ times clockwise, each by $5$ degrees, and then rotate back $12$ times anti-clockwise, by $5$ degrees each.
The experiment is presented in our video~(at 00:41).

In both tests we compared our system to one of the widely used techniques for pose estimation, which places a chessboard with known geometry in the observed scene.
The chessboard pattern is a basic representative for the family of fiducial marker-based approaches.
We used the same camera (PSEye3) for both methods.
A large A3 chessboard pattern was placed in front of the moving camera, such that it is clearly visible during the entire test.
Both methods aim to detect 2D features and find their corresponding 3D features in the predefined point cloud.
The PnP method (from OpenCV) was then used in both cases to recover the 6DoF~\cite{lepetit2009epnp}.

\iffalse
\begin{figure}[htbp]
\centering
\includegraphics[width=0.8\linewidth]{images/translation_test.png}
\caption{\textbf{Rails test position accuracy for each frame.} (Left:) 2D downwards view of the recovered camera position. (Right:) the recovered $Z$-axis values (camera height).}
\label{fig:RailTest}
\end{figure}

\begin{figure}[htbp]
\centering
\includegraphics[width=0.7\linewidth]{images/static_rotation_test.jpg}
\caption{\textbf{Rails test orientation accuracy } for each frame for each of the $3$ angles.}
\label{fig:RailsTest_angleErr}
\end{figure}
\fi

\begin{figure}[htbp]
    \centering
    \begin{subfigure}{\columnwidth}
        \includegraphics[width=\linewidth]{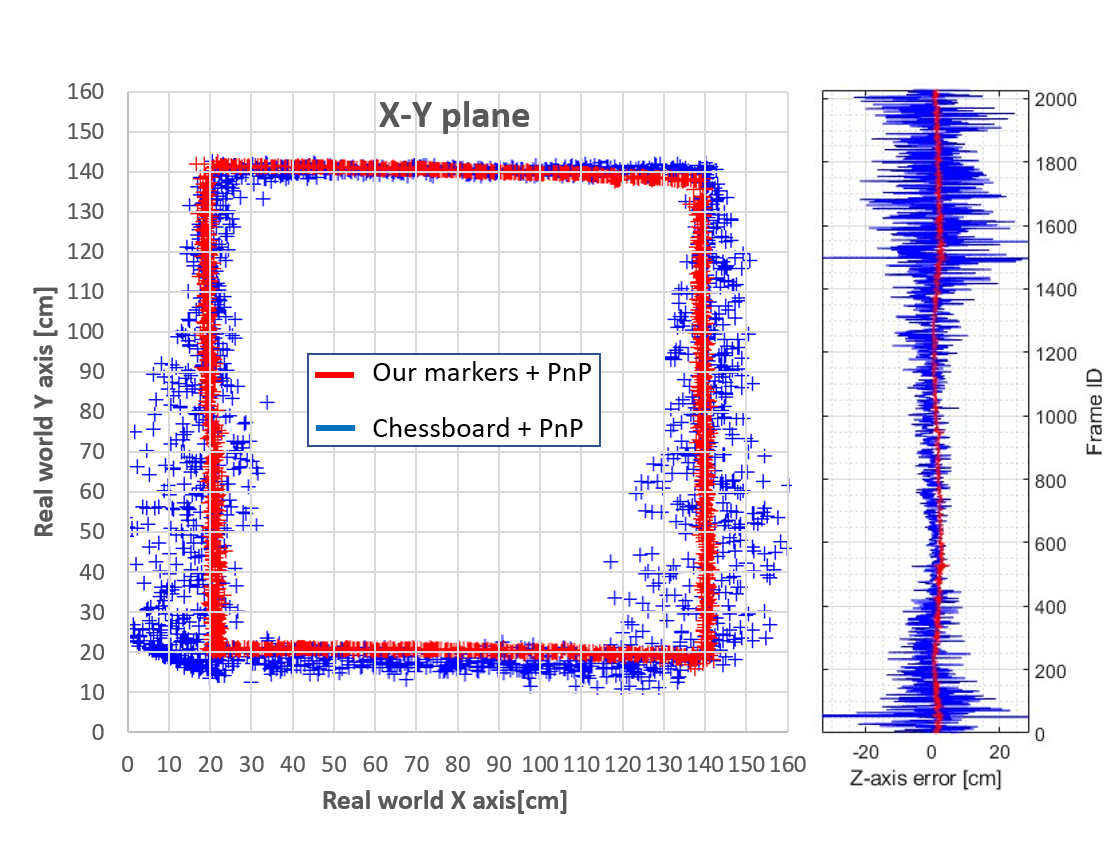}
        \caption{\textbf{Position accuracy.} (Left:) 2D top view of the recovered camera position. (Right:) the recovered $Z$-axis values (camera height).}
    \end{subfigure}
    \begin{subfigure}{\columnwidth}
        \includegraphics[width=\linewidth]{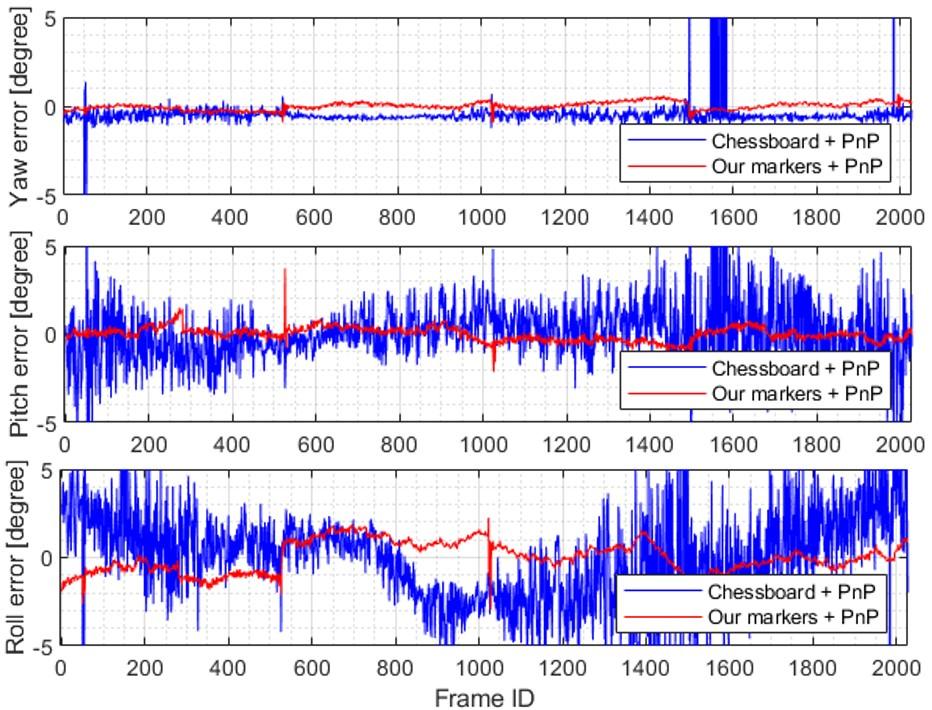}
        \caption{\textbf{Orientation accuracy }for each frame for each of the $3$ angles.}
    \end{subfigure}
    \caption{\textbf{Rail test accuracy.}}\label{fig:RailTest}
\end{figure}

\begin{figure}[htbp]
\centering
\includegraphics[width=\linewidth]{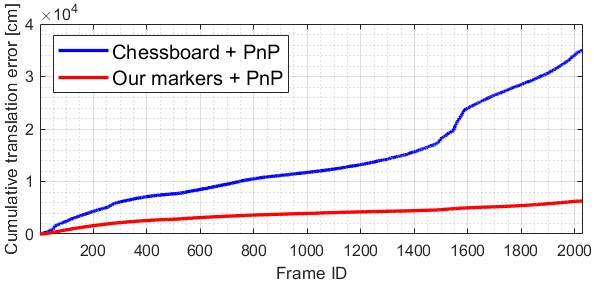}
\caption{\textbf{Rails test cumulative position error } compared to the ground truth, from the beginning of the test till each frame.
Using our markers, the total sum of errors throughout the entire experiment was roughly $0.6\cdot 10^4$, as compared to $3.5\cdot 10^4$ when using the chessboard pattern.}
\label{fig:RailsTest_cumulativeErr}
\end{figure}

\begin{figure}[htbp]
\centering
\includegraphics[width=\linewidth]{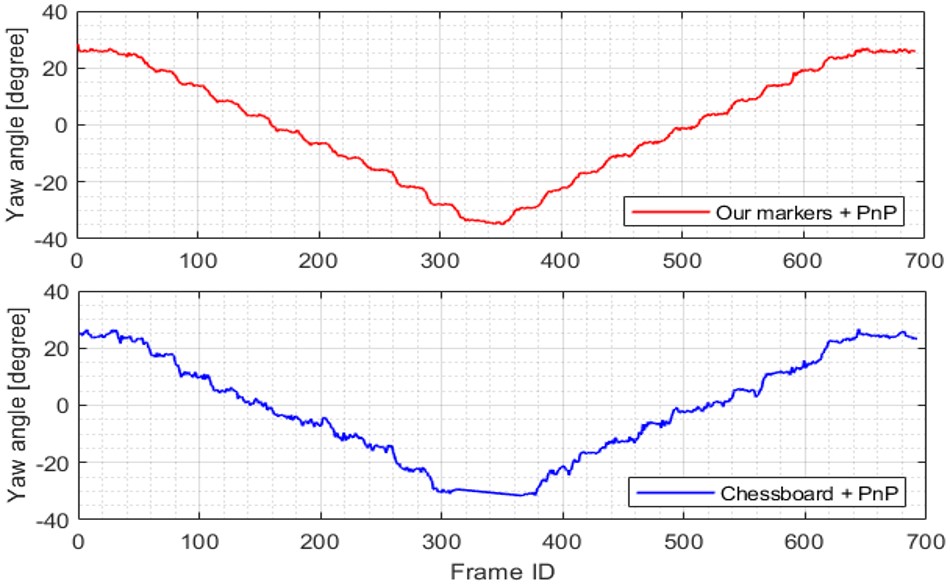}
\caption{\textbf{Servo test orientation accuracy and stability.}}
\label{fig:servo}
\end{figure}

Fig.~\ref{fig:RailTest}--\ref{fig:servo} present the accuracy results of both tests. As shown there, our system outperformed the widely used chessboard method, for both position and orientation accuracy.
Our system yielded an error up to $\times 5$ times smaller, while also being more stable and consistent.
Furthermore, our beacons deployment was easier to scale up in order to cover larger areas, and easily detected from far away compared to the chessboard; see our video~\cite{vid}.

\subsection{Handling fast moving objects and occlusions}
To test our system's performance as an external positional tracking system, we mounted our beacons on a fast moving toy race car, as illustrated in Fig.~\ref{Fig:BeaconDeploy}, and placed our perception unit in a fixed location in the room from which it can observe the race car.
The goal was to compute the 6DoF of the car in two different scenarios: (i) when the car is moving fast and/or performing complex maneuvers, and (ii) when the car passes behind some object and is occluded by it for some amount of time, before it is observed again.

The two tests are presented in the our video~\cite{vid} (at 01:09 and 01:54 respectively).
As shown, our system was able to accurately localize the fast moving car also when performing fast maneuvers.
It is also clear that our system requires a very small amount of time to re-track and re-localize the car after it has been occluded, as also predicted by Table~\ref{tab:tradeoff}.

\subsection{Challenging environments}
Many existing SLAM systems have very poor performance in environments which lack texture or are very repetitive, for example, a room with flat uniformly colored  walls, long symmetric hallways, and staircases.
To that end, we tested both the ORBSLAM2 system~\cite{mur2015orb}, which is a computationally-heavy and sophisticated system, and our system's performance in a typical two story staircase.
Unfortunately, the ORBSLAM2 system did not even manage to capture initial features in such an environment due to symmetry and uniform color.
Our system's performance, using the same optical sensor, is demonstrated in our video~\cite{vid} (at 02:26).
As presented, using $16$ of our coded beacons, we were able to accurately find the location and navigate in a stable manner within a two story staircase.
This test is a proof of concept that our low cost and computationally-light system can be integrated into an existing SLAM pipeline, such as ORBSLAM2, to help overcome challenging scenarios.

\subsection{Potential Use-Cases}
Other potential use-cases and applications for our system include:
\begin{enumerate}
\renewcommand{\labelenumi}{(\roman{enumi})}
    \item \textbf{Head tracking and pose estimation }of a human or a humanoid. This application is presented in our video~\cite{vid} (at 02:52).

    \item \textbf{Tracking and localization of multiple objects in the external approach.} Different beacons may be placed on more than one moving object, to obtain the 6DoF of multiple moving objects, e.g., a car and a drone simultaneously. However, this requires applying the initial mapping step for each object on its own.

    \item \textbf{Outdoor navigation }when either the lighting conditions are challenging e.g., at night or in a dynamic environment such as a forest or a plantation, where the plants are very similar and move rapidly. In this scenario, most vision (RGB) based methods will likely to be inaccurate and have a lot of outliers. However, our system is expected to work properly in the absence of light due to its use of static previously deployed IR-based beacons which are easily detected in such conditions.

    \item \textbf{Warehouse automation. }Using our coded beacons, manufacturing robots or inventory counting drones can autonomously localize and navigate accurately throughout the warehouse, even at dark.

    \item \textbf{Under water navigation }is a tough task~\cite{paull2013auv} mainly due to the lack of GPS signal, very poor lighting conditions, and highly noisy images which make the detection of image features an almost impossible task. This motivates the use of previously deployed markers such as our beacons, which simplifies the detection step in the noisy images, and are not affected by the lack of light. This can be useful for autonomous underwater research vessels.
\end{enumerate}
%\textbf{(i) Head tracking and pose estimation } as in our demonstration video~\cite{vid} (at 02:52).
%\textbf{(ii) Tracking and localization of multiple objects in the external approach.} Different beacons may be placed on more than a single moving object, to obtain the 6DoF of multiple moving objects, for example, a car and a drone simultaneously.
%\textbf{(iii) Outdoor navigation } when either the lighting conditions are challenging, or the environment is dynamic such as a plantation.
%Most vision (RGB) based methods would most likely  be inaccurate in such scenarios. However, our system is expected to work properly due to its use of static and previously deployed IR-based beacons which are easily detectable in such conditions.
%\textbf{(iv) Under water navigation } is a tough task~\cite{paull2013auv} mainly due to the lack of GPS signal, very poor lighting conditions, and highly noisy images which make the detection of image features an almost impossible task.
%This motivates the use of previously deployed markers such as our beacons, which simplifies the detection step in the noisy images.
%, and are not affected by the lack of light.

\section{Conclusions}
%In this paper we presented a beacon light coding protocol which is robust to noise and does not need direct signal communication.
%This protocol enables fast and accurate detection and identification of beacons with minimal computational burden with no need for signal communication or synchronization.
%We then show how to leverage such a protocol for the development of a cheap off-the-shelf positional tracking system, and introduced our own implementation.
%It can either operate as a stand-alone system or be integrated with a SLAM pipeline to enhance its performance.
%We tested our system's performance and compared it other common methods, and to ground truth, in different tasks.
%We believe that the proposed approach has high potential for increasing robustness in tracking for future industrial and possibly consumer level applications.

In this paper we presented a beacon light coding protocol which is robust to noise and does not need direct signal communication. This protocol enables fast and accurate detection and identification of beacons using an optical sensor with minimal computational burden.
We have explained in detail the development of our code-book and code words which are provably error resistant and have an efficient decoding process. We then provided a detailed description of the building blocks for a positional tracking system which is based on our protocol.
We continue by presenting an implementation of such a tracking system using low cost and accessible hardware, which can either operate as a stand-alone system or be integrated with a SLAM pipeline to enhance its performance.
The performance of this system was evaluated and compared to other common methods and to a ground truth.
Our system has low installation cost, low energy footprint, is easily deployable, can be easily adapted to different environments, does not need direct signal communication and can be used by multiple users without a centralized processing system. We believe this approach has much potential for increasing robustness in tracking for future industrial and possibly consumer level applications.


{\small
\bibliographystyle{ieee}
\bibliography{egbib}

\begin{thebibliography}{10}\itemsep=-1pt

\bibitem{opencode}
The authors commit to publish upon acceptance of this paper or reviewer
  request.

\bibitem{vid}
Demonstration video of our system and experiments:
  \url{{https://drive.google.com/file/d/1yD-4W4hqaIUhafevhDh9sYz2IfJK64dn/view?usp=sharing}}.

\bibitem{borg2005modern}
I.~Borg and P.~J. Groenen.
\newblock {\em Modern multidimensional scaling: Theory and applications}.
\newblock Springer Science \& Business Media, 2005.

\bibitem{opencv_library}
G.~Bradski.
\newblock {The OpenCV Library}.
\newblock {\em Dr. Dobb's Journal of Software Tools}, 2000.

\bibitem{bruckstein2000new}
A.~M. Bruckstein, R.~J. Holt, T.~S. Huang, and A.~N. Netravali.
\newblock New devices for 3d pose estimation: Mantis eyes, agam paintings,
  sundials, and other space fiducials.
\newblock {\em International Journal of Computer Vision}, 39(2):131--139, 2000.

\bibitem{diaz2019review}
E.~M. Diaz, D.~B. Ahmed, and S.~Kaiser.
\newblock A review of indoor localization methods based on inertial sensors.
\newblock In {\em Geographical and Fingerprinting Data to Create Systems for
  Indoor Positioning and Indoor/Outdoor Navigation}, pages 311--333. Elsevier,
  2019.

\bibitem{faigl2013low}
J.~Faigl, T.~Krajn{\'\i}k, J.~Chudoba, L.~P{\v{r}}eu{\v{c}}il, and M.~Saska.
\newblock Low-cost embedded system for relative localization in robotic swarms.
\newblock In {\em 2013 IEEE International Conference on Robotics and
  Automation}, pages 993--998. IEEE, 2013.

\bibitem{fiala2005artag}
M.~Fiala.
\newblock Artag, a fiducial marker system using digital techniques.
\newblock In {\em 2005 IEEE Computer Society Conference on Computer Vision and
  Pattern Recognition (CVPR'05)}, pages 590--596. IEEE, 2005.

\bibitem{garrido2014automatic}
S.~Garrido-Jurado, R.~Mu{\~n}oz-Salinas, F.~J. Madrid-Cuevas, and M.~J.
  Mar{\'\i}n-Jim{\'e}nez.
\newblock Automatic generation and detection of highly reliable fiducial
  markers under occlusion.
\newblock {\em Pattern Recognition}, 47(6):2280--2292, 2014.

\bibitem{HTCVIVE}
HTC-VIVE.
\newblock tracking system.
\newblock "\url{https://www.vive.com}".

\bibitem{kabsch1976solution}
W.~Kabsch.
\newblock A solution for the best rotation to relate two sets of vectors.
\newblock {\em Acta Crystallographica Section A: Crystal Physics, Diffraction,
  Theoretical and General Crystallography}, 1976.

\bibitem{lepetit2009epnp}
V.~Lepetit, F.~Moreno-Noguer, and P.~Fua.
\newblock Epnp: An accurate o (n) solution to the pnp problem.
\newblock {\em International journal of computer vision}, 81(2):155, 2009.

\bibitem{Levenshtein}
V.~I. Levenshtein.
\newblock Binary codes capable of correcting deletions, insertions and
  reversals.
\newblock In {\em Soviet physics doklady}, volume~10, page 707, 1966.

\bibitem{liu2007survey}
H.~Liu, H.~Darabi, P.~Banerjee, and J.~Liu.
\newblock Survey of wireless indoor positioning techniques and systems.
\newblock {\em IEEE Transactions on Systems, Man, and Cybernetics, Part C
  (Applications and Reviews)}, 37(6):1067--1080, 2007.

\bibitem{mazuryk1996virtual}
T.~Mazuryk and M.~Gervautz.
\newblock Virtual reality-history, applications, technology and future.
\newblock 1996.

\bibitem{more1978levenberg}
J.~J. Mor{\'e}.
\newblock The levenberg-marquardt algorithm: implementation and theory.
\newblock In {\em Numerical analysis}, pages 105--116. Springer, 1978.

\bibitem{munkres1957algorithms}
J.~Munkres.
\newblock Algorithms for the assignment and transportation problems.
\newblock {\em Journal of the society for industrial and applied mathematics},
  5(1):32--38, 1957.

\bibitem{munoz2020ucoslam}
R.~Mu{\~n}oz-Salinas and R.~Medina-Carnicer.
\newblock Ucoslam: Simultaneous localization and mapping by fusion of keypoints
  and squared planar markers.
\newblock {\em Pattern Recognition}, page 107193, 2020.

\bibitem{mur2015orb}
R.~Mur-Artal, J.~M.~M. Montiel, and J.~D. Tardos.
\newblock Orb-slam: a versatile and accurate monocular slam system.
\newblock {\em IEEE transactions on robotics}, 31(5):1147--1163, 2015.

\bibitem{nasser2015low}
S.~Nasser, I.~Jubran, and D.~Feldman.
\newblock Low-cost and faster tracking systems using core-sets for
  pose-estimation.
\newblock {\em CoRR abs/1511.09120}, 2015.

\bibitem{niehorster2017accuracy}
D.~C. Niehorster, L.~Li, and M.~Lappe.
\newblock The accuracy and precision of position and orientation tracking in
  the htc vive virtual reality system for scientific research.
\newblock {\em i-Perception}, 8(3):2041669517708205.

\bibitem{nister2004efficient}
D.~Nist{\'e}r.
\newblock An efficient solution to the five-point relative pose problem.
\newblock {\em IEEE transactions on pattern analysis and machine intelligence},
  26(6):756--770, 2004.

\bibitem{OptiTrack}
OptiTrack.
\newblock Motion capture systems.
\newblock "\url{https://www.optitrack.com}".

\bibitem{ozyecsil2017survey}
O.~{\"O}zye{\c{s}}il, V.~Voroninski, R.~Basri, and A.~Singer.
\newblock A survey of structure from motion*.
\newblock {\em Acta Numerica}, 26:305--364, 2017.

\bibitem{paull2013auv}
L.~Paull, S.~Saeedi, M.~Seto, and H.~Li.
\newblock Auv navigation and localization: A review.
\newblock {\em IEEE Journal of Oceanic Engineering}, 39(1):131--149, 2013.

\bibitem{pearson2003comma}
J.~Pearson.
\newblock Comma-free codes.
\newblock {\em Proc. 3rd Int. Wshop on Symmetry in Constraint Satisfaction
  Problems}, pages 161--167, 2003.

\bibitem{raspberry}
RaspberryPi.
\newblock singel board computer.
\newblock "\url{https://www.raspberrypi.org/}".

\bibitem{taketomi2017visual}
T.~Taketomi, H.~Uchiyama, and S.~Ikeda.
\newblock Visual slam algorithms: a survey from 2010 to 2016.
\newblock {\em IPSJ Transactions on Computer Vision and Applications}, 9(1):16,
  2017.

\bibitem{triggs1999bundle}
B.~Triggs, P.~F. McLauchlan, R.~I. Hartley, and A.~W. Fitzgibbon.
\newblock Bundle adjustment—a modern synthesis.
\newblock In {\em International workshop on vision algorithms}, pages 298--372.
  Springer, 1999.

\bibitem{tsai1984uniqueness}
R.~Y. Tsai and T.~S. Huang.
\newblock Uniqueness and estimation of three-dimensional motion parameters of
  rigid objects with curved surfaces.
\newblock {\em IEEE Transactions on pattern analysis and machine intelligence},
  (1):13--27, 1984.

\bibitem{lightlist}
USCG.
\newblock Light lists 2019.
\newblock "\url{http://www.navcen.uscg.gov/?pageName=lightlists}", 2019.

\bibitem{Vicon}
Vicon.
\newblock Motion systems.
\newblock "\url{https://www.vicon.com/}".

\bibitem{wahba1965least}
G.~Wahba.
\newblock A least squares estimate of satellite attitude.
\newblock {\em SIAM review}, 7(3):409--409, 1965.

\bibitem{wetzler2018tracking}
A.~Wetzler and R.~Kimmel.
\newblock Tracking using encoded beacons, May~24 2018.
\newblock US Patent App. 15/817,811.

\bibitem{zhang2000flexible}
Z.~Zhang.
\newblock A flexible new technique for camera calibration.
\newblock {\em IEEE Transactions on pattern analysis and machine intelligence},
  22(11):1330--1334, 2000.

\end{thebibliography}
}
\end{document}